\documentclass[11pt,letterpaper]{article}
\usepackage{emnlp2016}
\usepackage{times}
\usepackage{latexsym}
\usepackage{url}
\usepackage{breakurl}
\usepackage{graphicx}
\usepackage{amsmath}
\usepackage{stmaryrd}
\usepackage{listings}
\usepackage{booktabs}
\usepackage{mdframed}
\usepackage{forest}
\usepackage{multirow}

\usepackage[inference]{semantic}

\emnlpfinalcopy

\def\emnlppaperid{722}

\newcommand{\pred}[1]{\textsc{#1}}
\newcommand{\prog}[1]{\texttt{\small{#1}}}

\newcommand{\pchoose}[1]{\prog{choose}}
\newcommand{\pscore}[1]{\prog{score}}
\newcommand{\presolve}[1]{\prog{resolve}}
\newcommand{\true}[1]{\prog{true}}
\newcommand{\false}[1]{\prog{false}}
\newcommand{\unassigned}[1]{\prog{undef}}

\newcommand{\dataname}{\textsc{foodwebs}}
\newcommand{\pwbaseline}{\textsc{Worlds}}

\newcommand{\mycaption}[1]{{\renewcommand{\baselinestretch}{0.9} \caption{#1}}}

\title{Semantic Parsing to Probabilistic Programs for \\ Situated
  Question Answering}

\author{Jayant Krishnamurthy \and Oyvind Tafjord \and Aniruddha Kembhavi \\
  Allen Institute for Artificial Intelligence\\
  {\tt jayantk,oyvindt,anik@allenai.org} \\
}

\date{}

\begin{document}

\maketitle

\begin{abstract}
Situated question answering is the problem of answering questions
about an environment such as an image or diagram. This problem
requires jointly interpreting a question and an environment using
background knowledge to select the correct answer. We present Parsing
to Probabilistic Programs ($P^3$), a novel situated question answering
model that can use background knowledge and global features of the
question/environment interpretation while retaining efficient
approximate inference. Our key insight is to treat semantic parses as
probabilistic programs that execute nondeterministically and whose
possible executions represent environmental uncertainty. We evaluate
our approach on a new, publicly-released data set of 5000 science
diagram questions, outperforming several competitive classical and
neural baselines.

\end{abstract}


\section{Introduction}

\begin{figure}[th!]
  \small
  \begin{mdframed}[innerleftmargin=3pt,innertopmargin=3pt,innerbottommargin=3pt,innerrightmargin=3pt]
\begin{center}
  \includegraphics[width=2in]{figures/516.png} \\
\end{center}
\vspace{-.05in}
1. According to the given food chain, what is the number of organisms that eat deer?
(A) 3 (B) 2 (C) 4 \textbf{(D) 1} \\
~ \vspace{-.1in} \\
2. Which organism is both predator and prey?
(A) Bark Beetles \textbf{(B) Insect-eating birds} (C) Deer (D) Hawks \\
~ \vspace{-.1in} \\
3. Based on the given food web, what would happen if there were no insect-eating birds?
\textbf{(A) The grasshopper population would increase.} (B) The grasshopper population would decrease. (C) There would be no change in grasshopper number.
  \end{mdframed}
  \vspace{-.05in}
\mycaption{Example food web questions. A food web depicts a collection
  of organisms in an ecosystem with an arrow from organism $x$ to $y$
  indicating that $y$ eats $x$. Questions may require counting (1),
  knowing animal roles (2) and reasoning about population changes
  (3).}
\label{fig:example}
\vspace{-.15in}
\end{figure}


Situated question answering is a challenging problem that requires
reasoning about uncertain interpretations of both a question and an
environment together with background knowledge to determine the
answer. To illustrate these challenges, consider the 8th grade science
diagram questions in Figure \ref{fig:example}, which are motivated by
the Aristo project \cite{clark2016}. These questions require both
computer vision to interpret the diagram and compositional question
understanding. These components, being imperfect, introduce
uncertainty that must be jointly reasoned about to avoid implausible
interpretations. These uncertain interpretations must further be
combined with background knowledge, such as the definition of a
``predator,'' to determine the correct answer.

%

The challenges of situated question answering have not been completely
addressed by prior work. Early ``possible worlds'' models
\cite{matuszek2012,krishnamurthy2013jointly,malinowski2014multi} were
capable of compositional question understanding and using background
knowledge, but did not jointly reason about environment/question
uncertainty. These models also used unscalable inference algorithms
for reasoning about the environment, despite the lack of joint
reasoning. More recent neural models
\cite{antol2015,malinowski2015,yang2015} are incapable of using
background knowledge and it remains unclear to what extent these
models can represent compositionality in language.

We present Parsing to Probabilistic Programs ($P^3$), a novel approach
to situated question answering that addresses these challenges. It is
motivated by two observations: (1) situated question answering can be
formulated as semantic parsing with an execution model that is a
\emph{learned function of the environment}, and (2)
\emph{probabilistic programming} is a natural and powerful method for
specifying the space of permissible execution models and learning over
it. In $P^3$, we define a domain theory for the task as a probabilistic
program, then train a joint loglinear model to semantically parse
questions to logical forms in this theory and execute them in an
environment. Importantly, the model includes global features over
parsing and execution that enable it to avoid unlikely joint
configurations. $P^3$ leverages semantic parsing to represent
compositionality in language and probabilistic programming to specify
background knowledge and perform linear-time approximate inference
over the environment.


We present an experimental evaluation of $P^3$ on a new data set of
5000 food web diagram questions (Figure \ref{fig:example}). We compare
our approach to several baselines, including possible worlds and
neural network approaches, finding that $P^3$ outperforms both. An
ablation study demonstrates that global features help the model
achieve high accuracy. We also demonstrate that $P^3$ improves
accuracy on a previously published data set. Finally, we have released
our data and code to facilitate further research.

\section{Prior Work}

%

Situated question answering is often formulated in terms of parsing
both the question and environment into a common meaning representation
where they can be combined to select the answer. This general approach
has been implemented using different meaning representations:

\textbf{Possible world models} use a logical meaning representation
defined by a knowledge base schema. These models train a semantic
parser to map questions to queries and an environment model to map
environments to knowledge bases in this schema. Executing the queries
against the knowledge bases produces answers. These models assume that
the parser and environment model are independent and furthermore that
the knowledge base consists of independent predicate instances
\cite{matuszek2012,krishnamurthy2013jointly,malinowski2014multi}. Despite
these strong independence assumptions, these models have intractable
inference. An exception is Seo et al., \shortcite{seo2015} who
incorporate hard constraints on the joint question/environment
interpretation; however, this approach does not generalize to soft
constraints or arbitrary logical forms. In some work only the
environment model is learned
\cite{kollar2010,tellex2011,howard2014natural,howard2014efficient,berant2014modeling,krishnamurthy2015}.

\textbf{Neural networks} use a vector meaning representation that
encodes both the question and environment as vectors. These networks
have mostly been applied to visual question answering
\cite{antol2015}, where many architectures have been proposed
\cite{malinowski2015,yang2015,fukui2016}. It is unclear to what extent
these networks can represent compositionality in language using their
vector encodings. Dynamic Neural Module Networks
\cite{andreas2016deep,andreas2016} are the exception to the above
generalization. This approach constructs a neural network to represent
the meaning of the question via semantic parsing, then executes this
network against the image to produce an answer. Our approach is
similar except that we construct and execute a probabilistic
program. Advantages of our approach are that it naturally represents
the discrete structure of food webs and can use background knowledge.


\textbf{Preliminaries} for our work are semantic parsing and
probabilistic programming. Semantic parsing translates natural
language questions into executable logical forms and has been used in
applications such as question answering against a knowledge base
\cite{zelle1993,zettlemoyer05,liang11,kwiatkowski2013,berant2013,reddy2014large,yih2015,xu2016},
direction following \cite{chen_aaai11,artzi2013}, and information
extraction \cite{krishnamurthy2012,choi2015}. Semantic parsing alone
is insufficient for situated question answering because it does not
interpret the environment; many of the above approaches use semantic
parsing as a component in a larger model.

Probabilistic programming languages extend programming languages with
primitives for nondeterministic choice
\cite{mccarthy1963,goodman2014dippl}. We express logical forms in a
probabilistic variant of Scheme similar to Church \cite{goodman08};
however, this paper uses Python-like pseudocode for clarity. The
language has a single choice primitive called \pchoose{} that
nondeterministically returns one of its arguments. For example
\prog{choose(1,2,3)} can execute three ways, returning either 1, 2, or
3. Multiple calls to \pchoose{} can be combined. For example,
\prog{choose(1,2)+choose(1,2)} adds two nondeterministically chosen
values, and therefore has four executions that return 2, 3, 3 and 4.
Each execution also has a probability; in our case, these
probabilities are assigned by a trained model given the environment
and not explicitly specified in the program.


\section{Parsing to Probabilistic Programs ($P^3$)}

The $P^3$ model is motivated by two observations. The first is that
situated question answering can be formulated as semantic parsing with
an execution model that is a learned function of the
environment. Consider the first question in Figure
\ref{fig:example}. The meaning of this question could be represented
by a logical form such as $\pred{count}(\lambda x.\pred{eats}(x,
\pred{deer}))$, which we could train a semantic parser to predict
given a suitable \emph{domain theory} of functions such as
\pred{count} and \pred{eats}. However, the information required to
execute this logical form and answer the question must be extracted
from the diagram. Specifically, $\pred{eats}(x, y)$ depends on whether
an arrow is present between $x$ and $y$, which we must train a vision
model to determine. Thus, $\pred{eats}$ should be a \emph{learned
  function of the environment}.


This first observation suggests a need for a formalism for
representing uncertainty and performing learning over the domain
theory's functions. Our second observation is that probabilistic
programming is a natural fit for this task. In this paradigm, the
domain theory is a probabilistic program that defines the information
to be extracted from the environment by using \pchoose{}. To a first
approximation, the diagram question theory includes:

{
\lstset{language=Python, basicstyle=\ttfamily}
{\footnotesize
\begin{lstlisting}{}
def eats(x, y)
  choose(true, false)
\end{lstlisting}
}
}

Logical forms are then probabilistic programs, each of whose possible
executions represents a different interpretation of the
environment. For example, executing $\pred{eats}(\pred{lion},
\pred{deer})$ hits the \pchoose{} in the above definition, resulting
in two executions where the lion either eats or does not eat the
deer. In the \pred{count} example above, each execution represents a
different set of animals that eat the deer. To learn the correct
environment interpretation, we train an execution model to assign a
probability to each execution given features of the environment.
Using probabilistic programming enables us to combine learned
functions, such as \pred{eats}, with background knowledge functions,
such as \pred{count}, and also facilitates inference.


According to these observations, applying $P^3$ has two steps. The
first step is to define an appropriate domain theory. This theory is
the main design decision in instantiating $P^3$ and provides a
powerful way to encode domain knowledge. The second step is to train a
loglinear model consisting of a semantic parser and an execution
model. This model learns to semantically parse questions into logical
forms in the theory and execute them in the environment to answer
questions correctly.
We defer discussion of the diagram question domain theory to Section
\ref{sec:dqa} and focus on the loglinear model in this section.

\subsection{Model Overview}

The input to the $P^3$ model is a question and an environment and its
output is a denotation, which is a formal answer to the
question. $P^3$ is a loglinear model with two factors: a semantic
parser and an execution model. The semantic parser scores syntactic
parses and logical forms for the question. These logical forms are
probabilistic programs with multiple possible executions (specified by
the domain theory), each of which may return a different
denotation. The execution model assigns a score to each of these
executions given the environment. Formally, the model predicts a
denotation $\gamma$ for a question $q$ in an environment $v$ using
three latent variables:
\begin{align*}
  P(\gamma | v, q; \theta) = & \sum_{e, \ell, t} P(e, \ell, t | v, q; \theta) \textbf{1}(\text{ret}(e) = \gamma) \\
P(e, \ell, t | v, q; \theta) = & \frac{1}{Z_{q,v}} f_{\text{ex}}(e, \ell, v; \theta_{\text{ex}})f_{\text{p}}(\ell, t, q; \theta_{\text{p}}) 
\end{align*}

The model is composed of two factors. $f_{\text{p}}$ represents the
semantic parser that scores logical forms $\ell$ and syntactic parse
trees $t$ given question $q$ and parameters
$\theta_{\text{p}}$. $f_{\text{ex}}$ represents the execution
model. Given parameters $\theta_{\text{ex}}$, this factor assigns a
score to a logical form $\ell$ and its execution $e$ in environment
$v$. The denotation $\gamma$, i.e., the formal answer to the question,
is simply the value returned by $e$. $Z_{q,v}$ represents the model's
partition function. The following sections describe these factors in
more detail.

\subsection{Semantic Parser}

\begin{figure}[t]
{\scriptsize
\begin{center}
  \begin{minipage}[h]{\linewidth}
\setpremisesend{0pt}
\setpremisesspace{0pt}
\setnamespace{1pt}
\inference{
  \inference{
    \inference{\mbox{if}}{\begin{gathered}S/N/S : \vspace{-.05in}\\ \lambda x.\lambda y.\lambda f. \vspace{-.05in}\\ \pred{cause}(x, f(y)) \vspace{.032in} \end{gathered}} &
    \inference{
      \inference{\mbox{mice}}{\begin{gathered}N : \vspace{-.05in}\\ \pred{mice} \end{gathered}} &
      \inference{\mbox{die}}{\begin{gathered}S\backslash N : \vspace{-.05in}\\ \lambda x.\pred{decrease}(x) \end{gathered}}
    }{
      S : \pred{decrease}(\pred{mice})
    }
  }{
    S/N : \lambda y.\lambda f.\pred{cause}(\pred{decrease}(\pred{mice}), f(y))
  }  &
  \inference{\mbox{snakes}}{\begin{gathered}N : \vspace{-.05in}\\ \pred{snakes} \vspace{.34in} \end{gathered}} &
  \inference{\mbox{will \underline{\hspace{0.25cm}} ?}}{\begin{gathered} skip \\ \vspace{.29in} \end{gathered}}
}{
  S : \lambda f.\pred{cause}(\pred{decrease}(\pred{mice}), f(\pred{snakes}))
  }
\end{minipage}
\end{center}
}
\vspace{-.05in}
\mycaption{Example CCG parse of a question as predicted by the semantic
  parser $f_{\text{p}}$. The logical form $\ell$ for the question is shown
  on the bottom line.}
\label{fig:ccg-parse-example}
\vspace{-.12in}
\end{figure}

The factor $f_{\text{p}}$ represents a Combinatory Categorial
Grammar (CCG) semantic parser \cite{zettlemoyer05} that scores logical
forms for a question. Given a lexicon\footnote{In our experiments, we
  automatically learn the lexicon in a preprocessing step. See Section
  \ref{sec:baselinecomparison} for details.}  mapping words to syntactic
categories and logical forms, CCG defines a set of possible syntactic
parses $t$ and logical forms $\ell$ for a question $q$. Figure
\ref{fig:ccg-parse-example} shows an example CCG
parse. $f_{\text{p}}$ is a loglinear model over parses $(\ell, t)$:
\begin{align*}
f_{\text{p}}(\ell, t, q; \theta_{\text{p}}) = \exp \{ \theta_{\text{p}}^T \phi(\ell,t,q) \}
\end{align*}

The function $\phi$ maps parses to feature vectors. We use a rich set
of features similar to those for syntactic CCG parsing \cite{clark07};
a full description is provided in an online appendix.

\subsection{Execution Model}

The factor $f_{\text{ex}}$ is a loglinear model over the executions of
a logical form given an environment. Logical forms in $P^3$ are
probabilistic programs with a \emph{set} of possible executions, where
each execution $e$ is a sequence, $e = [e_0, e_1, e_2, ..., e_n]$.
$e_0$ is the program's starting state, $e_i$ represents the state
immediately after the $i$th call to \pchoose{}, and $e_n$ is the state
at termination. The score of an execution is:
\begin{align*}
f_{\text{ex}}(e, \ell, v; \theta_{\text{ex}}) & = \prod_{i=1}^{n} \exp \{ \theta^T_{\text{ex}} \phi(e_{i-1}, e_i, \ell, v) \}
\end{align*}

In the above equation, $\theta_{\text{ex}}$ represents the model's
parameters and $\phi$ represents a feature function that produces a
feature vector for the difference between sequential program states
$e_{i-1}$ and $e_i$ given environment $v$ and logical form
$\ell$. $\phi$ can include arbitrary features of the execution,
logical form and environment, which is important, for example, to
detect cycles in a food web (Section \ref{sec:execution-features}).

\subsection{Inference}
\label{sec:inference}


$P^3$ is designed to rely on approximate inference: our goal is to use
rich features to accurately make local decisions, as in linear-time
parsers \cite{nivre06malt}. We perform approximate inference using a
two-stage beam search. Given a question $q$, the first stage performs
a beam search over CCG parses to produce a list of logical forms
scored by $f_{\text{p}}$. This step is performed by using a CKY-style
chart parsing algorithm then marginalizing out the syntactic
parses. The second stage performs a beam search over executions of
each logical form. The space of possible executions of a logical form
is a tree (Figure \ref{fig:execution}) where each internal node
represents a partial execution up to a \pchoose{} call. The search
maintains a beam of partial executions at the same depth, and each
iteration advances the beam to the next depth, discarding the
lowest-scoring executions according to $f_{\text{ex}}$ to maintain a
fixed size beam. This procedure runs in time linear to the number of
\pchoose{} calls. We implement the search by rewriting the
probabilistic program into continuation-passing style, which allows
\pchoose{} to be implemented as a function that adds multiple
continuations to the search queue; we refer the reader to Goodman and
Stuhlm\"{u}ller \shortcite{goodman2014dippl} for details. Our
experiments use a beam size of 100 in the semantic parser, executing
each of the 10 highest-scoring logical forms with a beam of 100
executions.



\subsection{Training}

$P^3$ is trained by maximizing loglikelihood with stochastic gradient
ascent. The training data $\{(q^i, v^i, c^i)\}_{i=1}^n$ is a
collection of questions $q^i$ and environments $v^i$ paired with
\emph{supervision oracles} $c^i$. $c^i(e) = 1$ for a correct execution
$e$ and $c^i(e) = 0$ otherwise. The oracle $c^i$ can implement various
kinds of supervision, including: (1) labeled denotations, by verifying
the value returned by $e$ and (2) labeled environments, by verifying
each choice made by $e$. The oracle for diagram question answering
combines both forms of supervision (Section \ref{sec:dqa-training}).

The objective function $O$ is the loglikelihood of
predicting a correct execution:
\begin{align*}
  O(\theta) = & \sum_{i=1}^n \log \sum_{e,l,t} c^i(e) P(e, \ell, t | q^i, v^i; \theta)
\end{align*}

We optimize this objective function using stochastic gradient ascent,
using the approximate inference algorithm from Section
\ref{sec:inference} to estimate the necessary marginals. When
computing the marginal distribution over correct executions, we filter
each step of the beam search using the supervision oracle $c^i$ to
improve the approximation.

\section{Diagram Question Answering with $P^3$}
\label{sec:dqa}

As a case study, we apply $P^3$ to the task of answering food web
diagram questions from an 8th grade science domain. A few steps are
required to apply $P^3$. First, we create a domain theory of food webs
that represents extracted information from the diagram and background
knowledge for the domain. Second, we define the features of the
execution model that are used to learn how programs in the domain
theory execute given a diagram. Third, we define a component to select
a multiple-choice answer given a denotation. Finally, we define the
supervision oracle used for training.

\subsection{Food Web Diagram Questions}
\label{sec:fwdq}

We consider the task of answering food web diagram questions. The
input consists of a diagram depicting a food web, a natural language
question and a list of natural language answer options (Figure
\ref{fig:example}). The goal is to select the correct answer option.
This task has many regularities that require global features: for
example, food webs are usually acyclic and certain animals usually
have certain roles (e.g., mice are herbivores). We have collected and
released a data set for this task (Section \ref{sec:datacollection}).

We preprocess the diagrams in the data set using a computer vision
system that identifies candidate diagram elements
\cite{kembhavi2016}. This system extracts a collection of text labels
(via OCR), arrows, arrowheads and objects, each with corresponding
scores. It also extracts a collection of scored linkages between these
elements. These extractions are noisy and contain many discrepancies
such as overlapping text labels and spurious linkages. We use these
extractions to define a set of candidate organisms (using the text
labels), and also to define features of the execution model.

\subsection{Domain Theory}

%


The domain theory is a probabilistic program encoding the information
to extract from the environment as well as background knowledge about
food webs. It represents the structure of a food web using two
functions. These functions are predicates that invoke \pchoose{} to
return either true or false. The execution model learns to predict
which of these values is correct for each set of arguments given the
diagram. It furthermore has a collection of deterministic functions
that encode domain knowledge, including definitions of animal roles
such as \pred{herbivore} and a model of population change causation.

Figure \ref{fig:initialization-program} shows pseudocode for a portion
of the domain theory. Food webs are represented using two functions
over the extracted text labels: \pred{organism}$(x)$ indicates whether
the label $x$ is an organism (as opposed to, e.g., the diagram title);
and \pred{eats}$(x, y)$. The definitions of these functions invoke
\pchoose{} while remembering previously chosen values to avoid double
counting probabilities when executing logical forms such as
$\pred{organism}(\pred{deer}) \wedge
\pred{organism}(\pred{deer})$. The remembered values are stored in a
global variable that is also used to implement the supervision
oracle. Deterministic functions such as \pred{cause} are defined in
terms of these learned functions.

\begin{figure}[t]
\vspace{-.1in}
\lstset{language=Python, basicstyle=\ttfamily}
{\footnotesize
\begin{lstlisting}{}
# initialize predicate instance variables
# from text labels in environment
world = {"mice": undef,
         ("mice", "snakes"): undef, ...}
def organism(name)
  if (world[name] == undef)
    world[name] = choose(true, false)
  return world[name]

def eats(x, y)
  # same as organism but with pairs.

# entities referenced in the logical form
# must be organisms. choose() represents
# failure; it returns no values.
def getOrganism(x)
  if (organism(x)) return x else choose()

# change events are direction/
# text label tuples
def decrease(x)
  return ("decrease", x)
  
def cause(e1, e2)
  e12 = eats(e1[1], e2[1])
  e21 = eats(e2[1], e1[1])
  # deterministic model with cases. e.g.
  # if eats(y, x) then (cause (decrease x)
  # (decrease y)) -> true
  return doCause(e1[0], e2[0], e12, e21)
\end{lstlisting}
}
\vspace{-.08in}
\mycaption{Domain theory pseudocode for diagram question answering.
}
\vspace{-.15in}
\label{fig:initialization-program}
\end{figure}

The uses of \pchoose{} in the domain theory create a tree of possible
executions for every logical form. Figure \ref{fig:execution}
illustrates this tree for the logical form $\lambda
f.\pred{cause}(\pred{decrease}(\pred{mice}), f(\pred{snakes}))$, which
corresponds to the question ``what happens to the snakes when the mice
decrease?'' This logical form is shorthand for the following program:

\lstset{language=Python, basicstyle=\ttfamily}
{\footnotesize
\begin{lstlisting}{}
 filter(lambda f.cause(
     decrease(getOrganism("mice")),
     f(getOrganism("snakes"))),
   set(decrease, increase, unchanged))
\end{lstlisting}}

Specifically, entities such as \pred{mice} are created by calling
\prog{getOrganism} and logical forms with functional types implicitly
represent filters over the appropriate argument type. Executing this
program first applies the filter predicate to \prog{decrease}. Next,
it evaluates \prog{getOrganism("mice")}, which calls \prog{organism}
and encounters the first call to \pchoose{}. This call is shown as the
first branch of the tree in Figure \ref{fig:execution}. The successful
branch proceeds to evaluate \prog{getOrganism("snakes")}, shown as the
second branch. Finally, the successful branch evaluates \prog{cause},
which calls \prog{eats} twice, resulting in the final two
branches. The value returned by each branch is determined by the
causation model which performs some deterministic logic on the truth
values of the two eats relations.

\begin{figure}[t]
\centering
{\small
\begin{forest}
  for tree={l sep = 18pt,edge={->}}
  [{$\prog{organism(mice)}$},draw
    [ fail,edge label={node[midway,left] {false}} ]
    [ {$\prog{organism(snakes)}$},draw,edge label={node[midway,right] {true}}
      [ fail,edge label={node[midway,left] {false}} ]
      [ {$\prog{eats(mice,snakes)}$},draw,edge label={node[midway,right] {true}}
        [ {$\prog{eats(snakes,mice)}$},draw,edge label={node[midway,left] {false~}}
          [ {$\{\prog{unch.}\}$},edge label={node[midway,left] {false~}} ]
          [ {$\{\prog{dec.}\}$},edge label={node[midway,right] {~true}} ]
        ]
        [ {$\prog{eats(snakes,mice)}$},draw,edge label={node[midway,right] {~true}}
          [ {$\{\prog{inc.}\}$},edge label={node[midway,left] {false~}} ]
          [ {$\{\prog{}\}$},edge label={node[midway,right] {~true}} ]
        ]
      ]
    ]
  ]
\end{forest}
}
\vspace{-.25in}
\mycaption{Tree of possible executions for the logical form $\lambda
  f.\pred{cause}(\pred{decrease}(\pred{mice}),
  f(\pred{snakes}))$. Each path from root to leaf represents a single
  execution that returns the indicated denotation or fails, and each
  internal node represents a nondeterministic choice made with
  \pchoose{}.}
\label{fig:execution}
\vspace{-.15in}
\end{figure}

\subsection{Execution Features}
\label{sec:execution-features}

The execution model uses three sets of features: instance features,
predicate features, and denotation features. Instance features treat
each predicate instance independently, while the remainder are global
features of multiple predicate instances and the logical form. We
provide a complete listing of features in an online appendix.

\textbf{Instance features} fire whenever an execution chooses a truth
value for a predicate instance. These features are similar to the
per-predicate-instance features used in prior work to produce a
distribution over possible worlds. For \pred{organism}$(x)$, our
features are the vision model's extraction score for $x$ and indicator
features for the number of tokens in $x$. For \pred{eats}$(x,y)$, our
features are various combinations of the vision model's scores for
arrows that may connect the text labels $x$ and $y$.

\textbf{Predicate features} fire based on the global assignment of
truth values to all instances of a single predicate. The features for
\pred{organism} count occurrences of overlapping text labels among
true instances. The features for \pred{eats} include cycle count
features for various cycle lengths and arrow reuse features. The cycle
count features help the model learn that food webs are typically, but
not always, acyclic and the arrow reuse features aim to prevent the
model from predicting two different \pred{eats} instances on the basis
of a single arrow.

\textbf{Denotation features} fire on the return value of an
execution. There are two kinds of denotation features: size features
that count the number of entities in denotations of various types and
denotation element features for specific logical forms. The second
kind of feature can be used to learn that the denotation of $\lambda
x.\pred{herbivore}(x)$ is likely to contain \pred{mouse}, but unlikely
to contain \pred{wolf}.

\subsection{Answer Selection}
\label{sec:answer-selection}

$P^3$ predicts a distribution over denotations for each question,
which for our problem must be mapped to a distribution over multiple
choice answers. Answer selection performs this task using string match
heuristics and an LSTM \cite{hochreiter1997}. The string match
heuristics score each answer option given a denotation then select the
highest scoring answer, abstaining in the case of a tie. The score
computation depends on the denotation's type. If the denotation is a
set of entities, the score is an approximate count of the number of
entities in the denotation mentioned in the answer using a fuzzy
string match. If the denotation is a set of change events, the score
is a fuzzy match of both the change direction and the animal name. If
the denotation is a number, string matching is
straightforward. Applying these heuristics and marginalizing out
denotations yields a distribution over answer options.

A limitation of the above approach is that it does not directly
incorporate linguistic prior knowledge about likely answers. For
example, ``snake'' is usually a good answer to ``what eats mice?''
regardless of the diagram.  Such knowledge is known to be essential
for visual question answering \cite{antol2015,andreas2016} and
important in our task as well. We incorporate this knowledge in a
standard way, by training a neural network on question/answer pairs
(without the diagram) and combining its predictions with the string
match heuristics above. The network is a sequence LSTM that is applied
to the question concatenated with each answer option $a$ to produce a
50-dimensional vector $v_a$ for each answer. The distribution over
answers is the softmax of the inner product of these vectors with a
learned parameter vector $w$. For simplicity, we combine these two
components using a 50/50 mix of their answer distributions.

\subsection{Supervision Oracle}
\label{sec:dqa-training}

The supervision oracle for diagram question answering combines
supervision of both answers and environment interpretations. We assume
that each diagram has been labeled with a food web. An execution is
correct if and only if (1) all of the chosen values in the global
variable encoding the food web are consistent with the labeled food
web, and (2) string match answer selection applied to its denotation
chooses the correct answer. The first constraint guarantees that every
logical form has at most one correct execution for any given diagram.

\section{Evaluation}

Our evaluation compares $P^3$ to both possible worlds and neural
network approaches on our data set of food web diagram questions. An
ablation study demonstrates that both sets of global features improve
accuracy. Finally, we demonstrate $P^3$'s generality by applying it to
a previously-published data set, obtaining state-of-the-art results.

Code, data and supplementary material for this paper are available at:
{\small \burl{http://www.allenai.org/paper-appendix/emnlp2016-p3} }

\subsection{\dataname{} Data Set}
\label{sec:datacollection}

%

\dataname{} consists of $\sim$500 food web diagrams and $\sim$5000
questions designed to imitate actual questions encountered on 8th
grade science exams. The train/validation/test sets contain
$\sim$300/100/100 diagrams and their corresponding questions. The data
set has three kinds of annotations in addition to the correct answer
for each question. First, each diagram is annotated with the food web
that it depicts using \pred{organism} and \pred{eats}. Second, each
diagram has predictions from a vision system for various diagram
elements such as arrows and text labels \cite{kembhavi2016}. These are
noisy predictions, not ground truth. Finally, each question is
annotated by the authors with a logical form (or null if its meaning
is not representable in the domain theory). These logical forms are
not used to train $P^3$ but are useful to measure per-component error.

We collected \dataname{} by using a crowdsourcing process to expand a
collection of real exam questions. First, we collected 89 questions
from 4th and 8th grade exams and 500 food web diagrams using an image
search engine. Second, we generated questions for these diagrams using
Mechanical Turk. Workers were shown a diagram and a real question for
inspiration and asked to write a new question and its answer
options. We validated each generated question by asking 3 workers to
answer it, discarding questions where at least 2 did not choose the
correct answer. We also manually corrected any ambiguous (e.g., two
answer options are correct) and poorly-formatted (e.g., two answer
options have the same letter) questions. The final data set has high
quality: a human domain expert correctly answered 95 out of 100
randomly-sampled questions.


\subsection{Baseline Comparison}
\label{sec:baselinecomparison}

Our first experiment compares $P^3$ with several baselines for
situated question answering. The first baseline, \pwbaseline{}, is a
possible worlds model based on Malinowski and Fritz
\shortcite{malinowski2014multi}. This baseline learns a semantic
parser $P(\ell, t | q)$ and a distribution over food webs $P(w | v)$,
then evaluates $\ell$ on $w$ to produce a distribution over
denotations. This model is implemented by independently training
$P^3$'s CCG parser (on question/answer pairs and labeled food webs)
and a possible-worlds execution model (on labeled food webs). The CCG
lexicon for both $P^3$ and \pwbaseline{} was generated by applying
\textsc{PAL} \cite{krishnamurthy2016} to the same data. Both models
select answers as described in Section \ref{sec:answer-selection}.

We also compared $P^3$ to several neural network baselines. The first
baseline, \textsc{Lstm}, is the text-only answer selection model
described in Section \ref{sec:answer-selection}. The second baseline,
\textsc{Vqa}, is a neural network for visual question answering. This
model represents each image as a vector by using the final layer of a
pre-trained VGG19 model \cite{simonyan2014} and applying a single
fully-connected layer. It scores answer options by using the answer
selection LSTM to encode question/answer pairs, then computing a dot
product between the text and image vectors. This model is somewhat
limited because VGG features are unlikely to encode important diagram
structure, such as the content of text labels. Our third baseline,
\textsc{Dqa}, is a neural network that rectifies this limitation
\cite{kembhavi2016}. It encodes the diagram predictions from the
vision system as vectors and attends to them using the LSTM-encoded
question vector to select an answer. This model is trained with
question/answer pairs and diagram parses, which is roughly comparable
to the supervision used to train $P^3$.

\begin{table}
  \centering
  \small
  \begin{tabular}{lcc} \toprule
    & & Accuracy \\
    Model & Accuracy & (Unseen Organisms) \\ \midrule
    $P^3$ & \textbf{69.1} & \textbf{57.7} \\
    \pwbaseline{} & 63.6 & 50.8 \\
    \textsc{Lstm} & 60.3 & 34.7 \\
    \textsc{Vqa} & 56.5 & 36.8 \\
    \textsc{Dqa} & 59.3 & 33.0 \\ \midrule
    Random & 25.2 & 25.2 \\ \bottomrule
  \end{tabular}
  \mycaption{Accuracy of $P^3$ and several baselines on the \dataname{}
    test set and a modified test set with unseen organisms.}
  \vspace{-.15in}
  \label{table:neural}
\end{table}

\begin{table}
  \centering
  \small
  \begin{tabular}{lcc} \toprule
    Model & Accuracy & $\Delta$ \\ \midrule
  $P^3$ & 69.1 & \\
  ~ -LSTM & 59.8 & -9.3 \\
  ~ -LSTM -denotation & 55.8 & -13.3 \\
  ~ -LSTM -denotation -predicate & 52.4 & -16.7 \\ \bottomrule
  \end{tabular}


  \mycaption{Test set accuracy of $P^3$ removing LSTM answer selection
    (Section \ref{sec:answer-selection}), denotation features and
    predicate features (Section \ref{sec:execution-features}).}
  \vspace{-.15in}
  \label{table:ablation}
\end{table}


Table \ref{table:neural} compares the accuracy of $P^3$ to these
baselines. Accuracy is the fraction of questions answered
correctly. \textsc{Lstm} performs well on this data set, suggesting
that many questions can be answered without using the image. This
result is consistent with results on visual question answering
\cite{antol2015}. The other neural network models have similar
performance to \textsc{Lstm}, whereas both \pwbaseline{} and $P^3$
outperform it. We also find that $P^3$ outperforms \pwbaseline{}
likely due to its global features, which we investigate in the next
section.

Given these results, we hypothesized that the neural models were
largely memorizing common patterns in the text and were not able to
interpret the diagram. We tested this hypothesis by running each model
on a test set with unseen organisms created by reversing the organism
names in every question and diagram (Table \ref{table:neural}, right
column). As expected, the accuracy of \textsc{Lstm} is considerably
reduced on this data set. \textsc{Vqa} and \textsc{Dqa} again perform
similarly to \textsc{Lstm}, which is consistent with our
hypothesis. In contrast, we find that the accuracies of \pwbaseline{}
and $P^3$ are only slightly reduced, which is consistent with superior
diagram interpretation abilities but ineffective LSTM answer
selection.

\subsection{Ablation Study}
\label{sec:ablation}

We performed an ablation study to further understand the impact of
LSTM answer selection and global features. Table \ref{table:ablation}
shows the accuracy of $P^3$ trained without these components. We find
that LSTM answer selection improves accuracy by 9 points, as expected
due to the importance of linguistic prior knowledge. Global features
improve accuracy by 7 points, which is roughly comparable to the delta
between $P^3$ and \pwbaseline{} in Table \ref{table:neural}.

\subsection{Component Error Analysis}

Our third experiment analyses sources of error by training and
evaluating $P^3$ while providing the gold logical form, food web, or
both as input. Table \ref{table:component} shows the accuracy of these
three models. The final entry shows the maximum accuracy possible
given our domain theory and answer selection. The larger accuracy
improvement with gold food webs suggests that the execution model is
responsible for more error than semantic parsing, though both
components contribute.


\subsection{\textsc{scene} Experiments}

Our final experiment applies $P^3$ to the \textsc{scene} data set of
Krishnamurthy and Kollar \shortcite{krishnamurthy2013jointly}. In this
data set, the input is a natural language expression, such as ``blue
mug to the left of the monitor,'' and the output is the set of objects
in an image that the expression denotes. The images are annotated with
a bounding box for each candidate object. The data set includes a
domain theory that was automatically generated by creating a category
and/or relation per word based on its part of speech. It also includes
a CCG lexicon and image features. We use these resources, adding
predicate and denotation features.

Table \ref{table:scene} compares $P^3$ to prior work on
\textsc{scene}. The evaluation metric is exact match accuracy between
the predicted and labeled sets of objects. We consider three
supervision conditions: QA trains with question/answer pairs, QA+E
further includes labeled environments, and QA+E+LF further includes
labeled logical forms. We trained $P^3$ in the first two conditions,
while prior work trained in the first and third conditions.
\textsc{KK2013} is a possible worlds model with a max-margin training
objective. $P^3$ slightly outperforms in the QA condition and $P^3$
trained with labeled environments outperforms prior work trained with
additional logical form labels.


\begin{table}
  \centering
  \small
  \begin{tabular}{lll} \toprule
    Model & Accuracy & $\Delta$ \\ \midrule
    $P^3$ & 69.1 &  \\
    ~ + gold logical form & 75.1 & +6.0 \\ 
    ~ + gold food web & 82.3 & +13.2 \\
    ~ + both & 91.6 & +22.5 \\ \bottomrule
  \end{tabular}

  
  \mycaption{Accuracy of $P^3$ when trained and evaluated with labeled
    logical forms, food webs, or both.}
  \label{table:component}
  \vspace{-.15in}
\end{table}

\begin{table}
  \centering
  \small
  \begin{tabular}{lccc} \toprule
    & \multicolumn{3}{c}{Supervision} \\
    Model      & QA & QA+E & QA+E+LF \\ \midrule
    $P^3$ & 68 & 75 & -- \\
    \textsc{KK2013} & 67 & -- & 70 \\ \bottomrule
   \end{tabular}
   \mycaption{Accuracy on the \textsc{scene} data set. \textsc{KK2013} results
     are from Krishnamurthy and Kollar (2013).}
     \vspace{-.2in}
   \label{table:scene}
\end{table}


\section{Conclusion}

Parsing to Probabilistic Programs ($P^3$) is a novel model for
situated question answering that jointly reasons about question and
environment interpretations using background knowledge to produce
answers. $P^3$ uses a domain theory -- a probabilistic program -- to
define the information to be extracted from the environment and
background knowledge. A semantic parser maps questions to logical
forms in this theory, which are probabilistic programs whose possible
executions represent possible interpretations of the environment. An
execution model scores these executions given features of the
environment. Both the semantic parser and execution model are jointly
trained in a loglinear model, which thereby learns to both parse
questions and interpret environments. Importantly, the model includes
global features of the logical form and executions, which help the
model avoid implausible interpretations. We demonstrate $P^3$ on a
challenging new data set of 5000 science diagram questions, where it
outperforms several competitive baselines.



\section*{Acknowledgments}

We gratefully acknowledge Minjoon Seo, Mike Salvato and Eric Kolve for
their implementation help, Isaac Cowhey and Carissa Schoenick for
their help with the data, and Oren Etzioni, Peter Clark, Matt Gardner,
Hannaneh Hajishirzi, Mike Lewis, and Jonghyun Choi for their comments.

%

\bibliography{semantic_parsing}
\bibliographystyle{emnlp2016}

\end{document}


\maketitle

\section{Semantic Parser}

The CCG semantic parser used in $P^3$ has a rich set of features based
on those for syntactic CCG parsing. Its lexicon entries have four
fields:

\begin{enumerate}
\item \textbf{Word sequence} -- The words for which the entry can be used.
\item \textbf{Syntactic category} -- The CCG syntactic category with head-passing markup.
\item \textbf{Logical form}
\item \textbf{Predicate} -- a representation of the entry's semantics
  used to populate dependency structures. We create a unique predicate
  per logical form.
\end{enumerate}

All of the fields of these lexicon entries are automatically produced
by running PAL \cite{krishnamurthy2016}. The parser also dynamically
creates lexicon entries using the OCR text labels in the environment
in order to identify references to organisms in the diagram.

During parsing, each chart entry consists of a syntactic category, a
predicate $X_p$ and a word index $X_i$. The predicate and word index
represent the head word of the entry. Which word is the head is
determined from the head passing markup on syntactic categories. Chart
entries also include logical forms; however, the parser's features do
not apply to these directly, rather they depend on logical forms via
predicates.

The CCG parser's features are:

\begin{enumerate}
\item \textbf{Lexicon entry counts} -- A feature per lexicon entry that counts the number of times it is used in the parse.
\item \textbf{Word skip counts} -- A feature per word that counts the number of times it is skipped in the parse.
\item \textbf{Dependency features} -- The parser creates dependency
  structures during parsing whenever an argument to a syntactic
  category is filled by applying a combinator. There is a feature per
  dependency structure that counts its number of occurrences in the
  parse. The dependency structure consists of the syntactic category,
  its associated predicate, the argument number that was filled, and
  the predicate of the argument category.
\item \textbf{Dependency distance features} -- These features capture
  the number of intervening words between the two words connected by a
  dependency structure. There is a dependency distance feature per
  syntactic category, associated predicate, argument number, and the
  number of words between the two dependency structure elements. The
  number of words is histogrammed into 4 bins $0,1,2,\ge 3$.
\item \textbf{Combinator features} -- Combinators are applied to a
  left and right syntactic category to produce a parent syntactic
  category. There is a combinator feature that counts occurrences of
  every left/right/parent triple.
\item \textbf{Headed combinator features} -- Same as above, but
  additionally including the predicate of the parent.
\item \textbf{Root features} - A feature per syntactic category and
  predicate pair indicating whether the pair appears at the root of
  the parse.
\end{enumerate}

\section{Execution Model}

The execution model's features rely on the predictions of a vision
system. This system outputs a collection of scored diagram elements
and linkages:

\begin{enumerate}
\item \textbf{Blobs} -- Objects detected in the image, e.g., the
  drawing of a mouse.
\item \textbf{Text} -- Text detected in the image using OCR. Each text
  element has a text value as well as a bounding box.
\item \textbf{Arrows} -- Arrows in the image, including simple lines
  and more complex bitmap arrows.
\item \textbf{Arrowheads} -- The heads of arrows with an accompanying
  bounding box.
\item \textbf{Intraobject Labels} -- A linkage associating a text element as the label of a blob.
\item \textbf{Interobject Linkage} -- A three-way linkage between two
  elements mediated by an arrow. The two elements can be either blobs
  or text elements.
\end{enumerate}

The execution features are functions of text elements and pairs of
text elements. As a preprocessing step, we associate each text element
with a blob by computing a maximum matching using the intraobject
label scores. We also associate a best arrow with each pair of text
labels using the interobject linkage scores.

The instance features for \pred{organism} are:

\begin{enumerate}
\item \textbf{Text Score} -- OCR score from the vision system for the
  detection.
\item \textbf{Bias} -- Always 1
\item \textbf{Sub-bounding Box} -- 1 if is the text element is
  contained in a larger text element's bounding box.
\item \textbf{Number of words} -- Two indicator features if the number of words is $\ge4$ or $\ge 6$.
\end{enumerate}

The instance features for \pred{eats} are:

\begin{enumerate}
  \item \textbf{Self Link} -- 1 if the instance is of the form
    $\pred{eats}(x,x)$
  \item \textbf{No Link} -- indicator feature that is 1 if no
    interobject linkage connects the two text elements.
  \item \textbf{Link Exists} -- indicator feature that is 1 if an
    interobject linkage connects the two text elements, either
    directly or through their associated blobs.
  \item \textbf{Link Score} -- score of the interobject linkage
    connecting the two text labels. This score is both histogrammed
    into bins $[0, 0.05, 0.1, 0.2, 0.3, 0.4, 0.5, 0.6, 0.7, 1]$ and
    included as a real-valued feature.
  \item \textbf{Link Path Score} -- product of the interobject linkage
    score and any intraobject label scores. This score differs from
    the above score if the text labels are connected through blobs, as
    it also incorporates the intraobject label scores connecting the
    text labels and blobs. This score is both histogrammed into bins
    $[0, 0.01, 0.05, 0.1, 0.3, 0.7, 1.0]$ and included as a
    real-valued feature.
\end{enumerate}

The predicate features for \pred{eats} are:

\begin{enumerate}
\item \textbf{Cycle Features} -- These features approximately count
  the number of cycles of various lengths in the graph of true
  $\pred{eats}$ relation instances. There is a separate feature for
  each cycle length $k=2,3,4,\ge 5$. Let $n_k$ be the number of edges
  $(i,j)$ for which the shortest path from $j \rightarrow i$ has
  length $k-1$. The $k$th cycle feature's value is $n_k / k$. Note
  that if the graph has only one cycle and that cycle has length $k$,
  the value of the $k$th cycle detection feature is 1. In the $\ge 5$
  case, each edge completing a cycle of length $m$ adds $1/m$ to the
  feature's value.
\item \textbf{Arrow Reuse Features} -- We count for each arrow $a$ in
  the diagram the number of true eats relation instances for which $a$
  is the best arrow. The reuse features are count features for the
  number of arrows used $2,3,\ge4$ times.
\end{enumerate}

The denotation features are:

\begin{enumerate}
  \item \textbf{Denotation Size} by denotation type. For each type of
    logical form, a size feature for whether the denotation contains
    $0,1,2,3,\ge 4$ elements.
  \item \textbf{Count Value Features}. If the denotation is an integer, an
    indicator feature for each value in $0,1,2,3,\ge4$.
  \item \textbf{Role Features}. An indicator feature for each role and
    animal that occurs $\ge5$ times in the training set. These
    features fire if either the logical form returns the role of an
    animal or returns all animals with a specific role.
  \item \textbf{Direction Features}. An indicator feature for each
    change direction. These features fire if the denotation is a set
    of change directions.
\end{enumerate}

\bibliography{semantic_parsing}
\bibliographystyle{emnlp2016}